\documentclass[twoside,leqno,twocolumn]{article}

\usepackage[letterpaper]{geometry}

\usepackage{ltexpprt}
\usepackage{hyperref}

\usepackage{amsmath,amssymb,amsfonts}
\usepackage{algorithmic}
\usepackage{graphicx}
\usepackage{textcomp}
\usepackage{xcolor}

\usepackage{multirow}
\usepackage{marvosym}
\usepackage{subcaption}
\usepackage{booktabs}
\usepackage{makecell}
\usepackage{float}
\usepackage{microtype}
\usepackage{wrapfig}

\begin{document}

\newcommand\relatedversion{}
\renewcommand\relatedversion{\thanks{The full version of the paper can be accessed at \protect\url{https://arxiv.org/abs/1902.09310}}} 

\title{Measuring disentangled generative spatio-temporal representation}
\author{Sichen Zhao
\thanks{s3802901@student.rmit.edu.au, School of Computing Technologies, RMIT University, Melbourne, 3000, VIC, Australia}
\and Wei Shao
\thanks{wei.shao@rmit.edu.au, School of Computing Technologies, RMIT University, Melbourne, 3000, VIC, Australia}
\and Jeffrey Chan
\thanks{jeffrey.chan@rmit.edu.au, School of Computing Technologies, RMIT University, Melbourne, 3000, VIC, Australia}
\and Flora D. Salim
\thanks{flora.salim@rmit.edu.au, School of Computing Technologies, RMIT University, RMIT University, 3000, VIC, Australia}
}

\date{}

\maketitle


\fancyfoot[R]{\scriptsize{Copyright \textcopyright\ 20XX by SIAM\\
Unauthorized reproduction of this article is prohibited}}





\begin{abstract}Disentangled representation learning offers useful properties such as dimension reduction and interpretability, which are essential to modern deep learning approaches. 
  Although deep learning techniques have been widely applied to spatio-temporal data mining, there has been little attention to further disentangle the latent features and understanding their contribution to the model performance, particularly their mutual information and correlation across features.
  In this study, we adopt two state-of-the-art disentangled representation learning methods and apply them to three large-scale public spatio-temporal datasets. To evaluate their performance, we propose an internal evaluation metric focusing on the degree of correlations among latent variables of the learned representations and the prediction performance of the downstream tasks. Empirical results show that our modified method can learn disentangled representations that achieve the same level of performance as existing state-of-the-art ST deep learning methods in a spatio-temporal sequence forecasting problem. Additionally, we find that our methods can be used to discover real-world spatial-temporal semantics to describe the variables in the learned representation.

\end{abstract}

\section{Introduction}
\label{intro}
Representation learning \cite{bengio2013representation} can solve many fundamental problems in deep learning such as noise, redundancy, and the curse of dimensionality \cite{su2018learning}. Disentangled representation learning \cite{bengio2013representation,multivae,betavae,factorVAE} is an unsupervised learning technique that disentangles features of a target distribution into narrowly defined variables and keeps them as independent as possible. In comparison to the traditional methods such as autoencoder \cite{kramer1991nonlinear, Goodfellow-et-al-2016} and PCA (Principal Component Analysis), the disentangled representation can offer insight about the relationships of the latent features with semantic attributes \cite{betavae} and can provide additional benefits for the downstream tasks \cite{henaff2019data,creager2019flexibly}.

\begin{figure}[ht]
\begin{center}
    \includegraphics[width=0.4\textwidth]{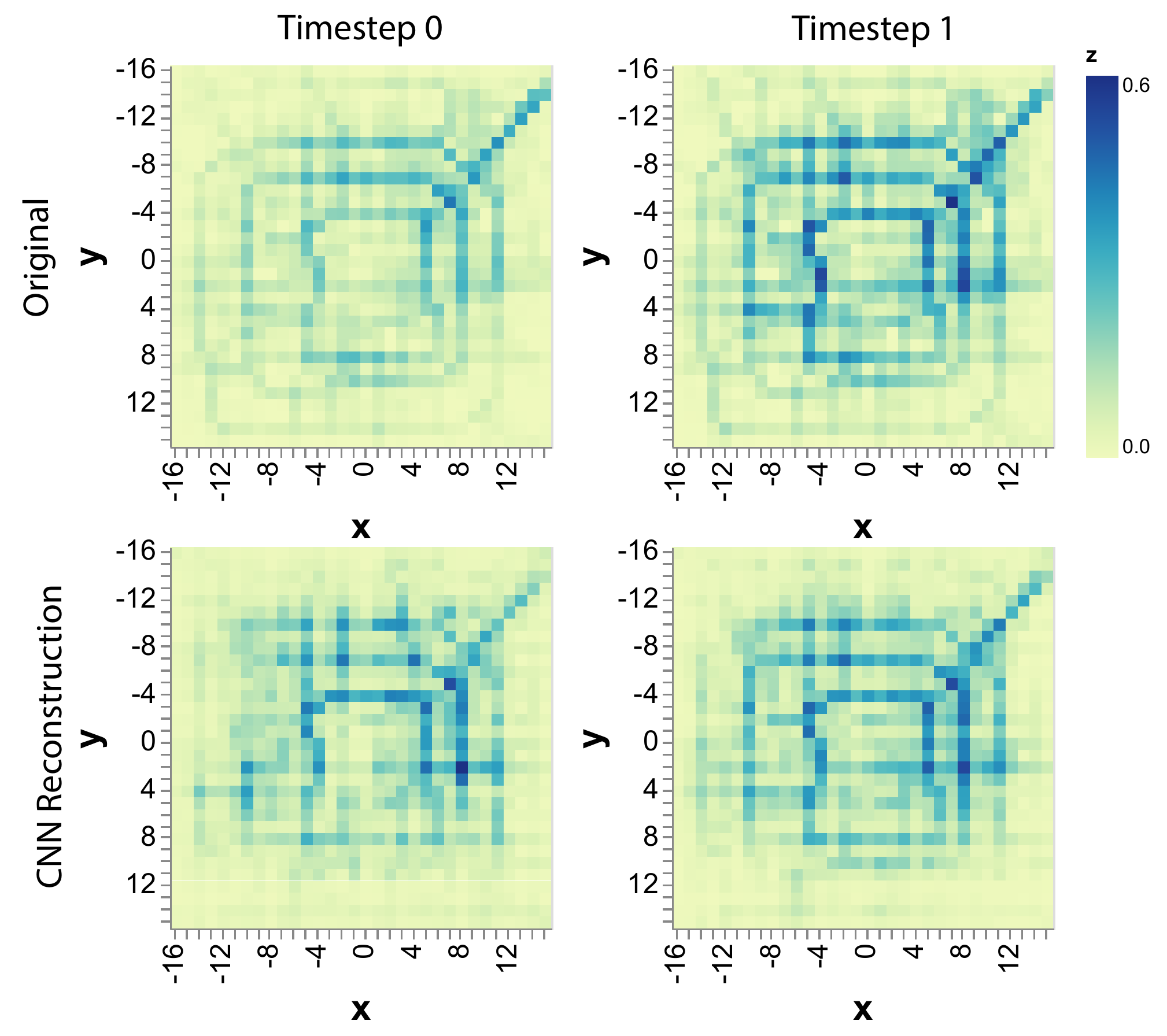}
\end{center}
\caption{Does treating the ST data as "image" lead to good disentangled representation? Reconstruction results on "compressed" image traffic flows in Beijing \cite{taxibj} using FactorVAE \cite{factorVAE}. The first row is the timestep 0-1 of the input image. The second row shows the the timestep 0-1 reconstruction using disentangled representation learning methods for image data.}
\label{image_recon}
\end{figure}

Many prior works on disentangled representation learning have shown promising results \cite{betavae,factorVAE,chen2018isolating}. Although representation learning can be applied to various data, most existing disentangled representation learning methods are developed with image data in mind. Thus, the common structure for decoder and encoder used in this area is CNNs (Convolutional Neural Networks). There is inadequate attention to disentangled representation learning for spatial-temporal (ST) data. Although ST data are regarded as image data in most existing research and are dealt with using similar approaches in the computer vision field, disentangled representation learning of ST data is not trivial to be solved. 
Existing disentangled representation learning methods are designed to extract information from the spatial domains and therefore struggle with information from the temporal domain, which is critical for ST data. As shown in Figure \ref{image_recon}, we 'compress' the spatio-temporal data into an image by setting each timestep as a single channel and applying representation learning methods to reconstruct the transformed image. 
Although the image model manages to preserve some network structure information, the difference across the different timesteps can hardly be identified, indicating only limited information is extracted from the temporal domain. Therefore, the first question is \textbf{"Is there a disentangled representation learning approach that can extract both spatial and temporal information of ST data?"}

One major challenge is to evaluate the quality of the data representations, given the difficulty in directly applying the current evaluation metrics on ST data. One widely-used evaluation metric in disentangled representation learning of image data is FactorVAE \cite{factorVAE}. First, they randomly pick a set of input $X$ with the fixed factor $f_{k}$ and extract the latent representation $z$ from each input $x$ in this set $X$. Then, multiple classifiers use this set of z as input to predict the index of that shared factor under a majority-vote scheme. The loss of this group of classifiers is used to represent the disentanglement of the representation. However, it demands the dataset to have a large number of informative labels, which are not generally available in ST datasets, since the real ST dataset might not have that many labels, and it is hard to match the data with the universal set of its labels. 
On the other hand, the current evaluation metrics only consider the disentanglement of the learned representations without paying attention to the possible downstream usage of the representations. One well-known hypothesis in this area assumes that a good disentangled representation is useful for the downstream tasks\cite{locatello2019challenging}, which is not true in many cases. 
Although some research shows great results by applying disentanglement methods on various downstream tasks, one recent work shows that the downstream tasks are not guaranteed to benefit from a representation with high disentanglement score \cite{locatello2019challenging}. Since the major goal for representation learning on ST data is associated with the downstream utility, it is crucial to find the relating factors behind the disentanglement and downstream task performance. Therefore, the second question of this work is: \textbf{"Is there a general way to evaluate an effective disentangled representation of ST data?"}

To address the aforementioned problems, we propose a novel framework to learn the disentangled representations of spatio-temporal data and evaluate the quality of learned representations. Our key contributions can be summarised as follows:
\begin{enumerate}
    \item We introduce a novel general evaluation metric for effective disentangled representation learning of spatio-temporal data and show how this metric can be estimated based on the downstream task and the observation data.
    \item We propose a novel approach, ST-VAE, to learn the disentangled representations of spatio-temporal data by introducing neural networks that are designed especially for the corresponding type of spatio-temporal data.
    \item We conduct several experiments on multiple benchmark spatio-temporal datasets in different categories to show that our methods can learn better representations. Meanwhile, the new representation is effective for downstream tasks. 
        We also investigate further potential usage of our approach in other directions, such as few-shot learning.
\end{enumerate}

\section{Related Work}
Most prior works on the disentangled representation learning problem are developed based on Variational Autoencoders (VAE) \cite{vae}, which is an unsupervised generative learning method. $\beta$-VAE, proposed by Higgins et al. \cite{betavae}, forces the inference model to disentangle the latent representation by adding a new hyperparameter $\beta$ to create an information bottleneck on the prior. FactorVAE \cite{factorVAE} further breaks down the objective function and tries to enhance disentanglement by penalising the total correlation of the learned representation. $\beta$-TCVAE \cite{chen2018isolating} also applies constraint on the total correlation but approximates it by estimating it for each mini-batch. 

The models mentioned above only focus on the disentanglement of image data, while disentangled representation learning of sequence data has less attention. S3VAE \cite{zhu2020s3vae} is proposed to separate static and dynamic factors of sequential data. Another approach proposed by \cite{li2018disentangled} is also focusing on separating the dynamic factors from static factors. Although they share a similar idea which uses an RNN-based architecture to extract dynamic factors for each timestep, 
they use different prior setups. Each frame in \cite{li2018disentangled} has its own content features while the time-invariant variables are shared by the whole sequence in S3VAE. As for the spatial-temporal type of data like video, SV2P \cite{babaeizadeh2017stochastic} uses the variational model to extract the time-invariant latent and make predictions for multiple frames. Models like \cite{denton2017unsupervised, hsieh2018learning} try to factorise each frame into a stationary part and a temporally dynamic component.

\section{Preliminaries}
There is no canonical definition for disentangled representation, one of the most popular definitions is from Bengio et al. \cite{bengio2013representation}: \textit{"a representation where a change in one dimension corresponds to a change in one factor of variation while being relatively invariant to changes in other factors."} This definition already implies several advantages of disentangled representation learning compared to the normal feature engineering/learning in classical machine learning approaches. Some also argue against adding interpretability into the definition of disentangled representation learning. Although it is not the case that every dimension can be linked to a real-world semantic factor, we can learn some insights by doing traversal on the latent only. 

Most of the disentangled representation learning methods are variants of Variational autoencoder(VAE). The original marginal likelihood of VAE is shown in Eq.\ref{eq:originall_vae}, and due to the intractability of the posterior distribution, it is not likely to calculate it directly. However, the author provided a variational lower bound (Evidence Lower Bound, ELBO), which is shown in Eq.\ref{eq:ELBO} can be optimised using stochastic gradient descent.

\begin{align}
log\ p_{\theta } (x) = &D_{KL} (q_{\phi } (z|x)||p_{\theta } (z))+\mathcal{L}_{VAE} (\theta ,\phi ;x) \label{eq:originall_vae} \\
\mathcal{L} (\theta ,\phi ;x) = &E_{q_{\phi } (z|x)} [log\ p_{\theta } (x|z)] \notag\\
 &-D_{KL} (q_{\phi } (z|x)||p_{\theta } (x))   \label{eq:ELBO}    
\end{align}

The first part in Eq.\ref{eq:ELBO} represents the \textit{reconstruction loss}, and by minimising this term, it forces the model to generate reliable reconstructions of the input and the synthesis data with better quality. The second term is a regularisation term for the posterior $q_{\phi } (z|x)$. 

$\beta$-VAE \cite{betavae} is the first model for introducing the disentanglement. By adding a hyper-parameter $\beta$ for penalising the Kullback-Liebler divergence term harder, the representation tends to become disentangled. Although the hyperparameter $\beta$ emphasises disentanglement, it suppresses the model's ability to produce a high-quality reconstruction. Based on that, the KL regularisation term are decomposed into the following form:

\begin{align}
D_{KL} (q_{\phi } (z|x)||p_{\theta } (z)) & =I( x,z) \notag\\
& =D_{KL}\left( q( z) ||\prod _{j=1} q( z_{j})\right) \notag\\
 & +\sum _{d} D_{KL}( q_{\phi }( z_{j}) ||p( z_{j})) ,z\in \mathbb{R}^{d}
 \label{eq:kl_decomp}
\end{align}

The first term describes the \textit{mutual information (MI)} between the input and its latent representation. The second term evaluates the level of dependency or redundancy among variables in the set evaluates the level of dependency or redundancy among variables in the set \cite{alfonso2010optimization} and is referred to as the \textit{Total Correlation (TC)} \cite{chen2018isolating}. Experimental results from $\beta$-TCVAE \cite{chen2018isolating} and FactorVAE \cite{factorVAE} show that, by amplifying the penalty on this term, the dependence between the variables is reduced hence emphasising the disentanglement. The third term in Eq.\ref{eq:kl_decomp} is defined as \textit{dimension-wise KL divergence}, which puts constraints on the generated latent code $z$ and pushes them towards their predefined Gaussian prior \cite{li2020pri}. We provide a summary of objectives typically observed in this problem in Table \ref{tab:ELBO_decompose}.

\begin{table*}[ht]
    \begin{center}
    \caption{The Objective decomposition of various disentangled VAEs.}
    \label{tab:ELBO_decompose}
    \begin{tabular}{c|cccc}
    \hline \hline 
     Methods & \makecell[c]{Recon\\loss} & \makecell[c]{Mutual\\Information} & \makecell[c]{Total\\Correlation} & \makecell[c]{Dimension\\-wise KL} \\
    \hline
    VAE\cite{vae} & \multirow{5}{*}{1} & 1 & 1 & 1 \\
    $\beta$-VAE\cite{betavae} &   & $\displaystyle \beta $ & $\displaystyle \beta $ & $\displaystyle \beta $ \\
    $\displaystyle \beta $-TCVAE\cite{chen2018isolating} &   & 1 & $\displaystyle \beta $ & 1 \\
    FactorVAE\cite{factorVAE} &   & 1 & $\displaystyle \gamma $ & 1 \\
     \hline \hline
    \end{tabular}
    \end{center}
\end{table*}

\section{Effective disentangled representation learning for Spatio-temporal data}
There are two major problems regarding learning a disentangled representation of spatio-temporal data: 1) There is no framework that can sufficiently extract disentangled representation from both spatial and temporal domains for real ST data. 2) It lacks a consistent approach to evaluate the model's performance with respect to spatio-temporal data. In this section, we discuss the difficulties in applying the disentangled representation learning for ST data and introduce our proposed metric.

\subsection{Evaluation of the disentangled representation }
The most recent works on disentangled representation learning are performed on images datasets such as dSprites \cite{dsprites17}, CelebA \cite{liu2015faceattributes} and 3DChairs \cite{aubry2014seeing}. Although these methods can be applied to spatio-temporal data with minor tweaks, this area still attracts not much attention due to the difficulty of proposing an evaluation metric of disentangled representation of the spatio-temporal data.

One of the most commonly used approaches to evaluate the performance of a disentangled representation learning or any generative model is by examining the quality of its synthetic results directly \cite{betavae,factorVAE}. Image data has a tremendous advantage here since it is fairly easy to visualise the synthesis image data, and human beings can easily detect the anomalies in the images. By contrast, it is difficult to recognise the anomalies in spatio-temporal data with common visualisation methods. Meanwhile, unlike image data, it is almost impossible to distinguish the original ST data and generative one by observing the visualisation results. Given the various forms of ST data, the visualisation needs to be adaptive, which makes it harder to produce and interpret in many cases.

Another way of evaluating the model's performance is by introducing certain metrics targeting the desired properties \cite{betavae,factorVAE,chen2018isolating}. BetaVAE\cite{betavae} and FactorVAE\cite{factorVAE} proposed metrics that quantifying disentanglement with one or multiple classifiers. However, those metrics focus on the relation between the real world factors and the learned representation. They require the datasets have a substantial amount of labels, which can be found in many image datasets, but not in ST data. . Although substantial efforts have been made to measure the disentanglement of the learned representation, the applicability is an issue due to the aforementioned problems.

Due to the aforementioned difficulties, a rigorous evaluation metric is needed to assess the performance of such representation formally. In this study, we adopted the definition of disentangled representation proposed by Bengio et al.\cite{bengio2013representation} and refined it based on the nature of ST data: "\textit{A representation that is informative enough for the downstream task while being as efficient as possible.}" This definition suggests two major properties of representation for ST data: \textbf{effectiveness} and \textbf{disentanglement}.

\subsubsection{Effective representation}
Effectiveness is an important property needed by the representation of ST data since the utility is a major goal for spatio-temporal deep learning tasks. Therefore, the introduction of disentangled representation learning should not compromise its downstream task utility. Besides, the effectiveness indicates the model's ability to preserve info from spatial and temporal domains. In this study, we formulate the effectiveness of representation of ST data in two parts: \textbf{reconstruction loss} and \textbf{downstream task utility}.

\textbf{Reconstruction loss: }  
The reconstruction loss is defined as follows:
\begin{equation}
    \text{ReconLoss} = -E_{q_{\phi }( z|x)}[ log\ p_{\theta }( x|z)]
\end{equation}
For spatial-temporal data, a high reconstruction loss either could indicate the model cannot extract info from both domains or lacks the ability to generate good synthetic ST data, like what we shows in Figure \ref{image_recon}.
 
\textbf{Downstream task utility: }
\label{sec:downstream}
There is an assumption that the learned representation can lead to better performance since it extracts most of the essential information into a much lower dimension \cite{bengio2013representation}. Although there are various types of tasks based on spatio-temporal data, we consider the simplest mobility prediction task in this work. The goal of this mobility prediction task is to predict the next several time frame given a spatio-temporal observation $S_c$. The predictive process can be formulated as: 
$\hat{s^{t+1}_{c}} =f_{r}\left( s^{0}_{c} ,s^{1}_{c} ,...,s^{t}_{c} ;\phi _{r}\right)$,  
where $\phi _{r}$ is the hyperparameters for the regressor. We use Root Mean Square Error (RMSE) as the loss of this task, and it can be formulated as follows: $\text{RMSE}=\sqrt{\frac{\sum _{c}\left( s^{t+1}_{c} \ -\hat{\ s^{t+1}_{c}}\right)^{2}}{C}}$, where $C$ is the total number of points (nodes) in each input. This term implies the performance of the learned representation in terms of its utility. If this term ends up getting very high, no matter how disentangled within the representation variables, whether it is useful for the downstream task is questioned.
\subsubsection{Disentangled representation}
Due to the difficulties in acquiring informative factors for the spatio-temporal dataset, most of the previously proposed metrics are hard to apply. To this end, we decide to use the value of \textbf{Mutual Information} minus the value of \textbf{Total Correlation} to indicate the disentanglement score of a latent representation. As shown in Table \ref{tab:ELBO_decompose}, prior works put regulations on both terms to enforce disentanglement, and the computation of those two terms does not demand informative factors.

\textbf{Index-Code Mutual Information: }
In this work, we estimate the index-code mutual information using the empirical mutual information estimation proposed by $\beta$-TCVAE \cite{chen2018isolating}. It defines that $q( z|n) =q( z|x_{n})$ and $q( z,n) =q( z|n) p( n) =q( z|n)\frac{1}{N}$, where $n\in\{ 1,2,...,N\}$ is the unique index that relates to each data point. Then, we use the joint distribution of a latent variable $z_j$ and a ground truth factor $v_k$ to estimate the mutual information under the following equation:

\begin{equation}
I( z_{j} ;v_{k}) =\mathbb{E}_{q( z_{j} ,v_{k})}\left[ log\sum _{n\in X} q( z_{j} |n) p( n|v_{k})\right] +H( z_{j})
\end{equation}

with the joint distribution $q( z_{j} ,v_{k})$ defined as follows:
$q( z_{j} ,v_{k}) =\sum ^{N}_{n=1} p( v_{k}) p( n|v_{k}) q( z_{j} |n)$. 
We perform the Minibatch Weighted Sampling (MWS) \cite{chen2018isolating} over the entire dataset instead of stratified sampling since this sampling mechanism is more generalised regardless of the underlying distribution.

The approximation value has highly relied on the estimation of the joint distribution $p(x, z)$. Prior works suggest penalising this term to get better disentanglement \cite{achille1706emergence}, while some works argue that a small value of $I(x, z)$ will impact the model's ability to reconstruct its input \cite{li2020pri}. 


\textbf{Total Correlation: }  
As first mentioned in $\beta$-TCVAE \cite{chen2018isolating}, the total correlation is a term that indicates the independence between latent variables. By emphasising the penalty on this term \cite{chen2018isolating,factorVAE,betavae}, the model is forced to make the latent variables statically independent, which leads to disentanglement. The total correlation can be calculated by the following equation:
\begin{align}
TC( z) & =D_{KL}( p( z) ||p( z_{1}) ...p( z_{d})) & \notag\\
 & =\int ...\int p( z) log\left(\frac{p( z)}{p( z_{1}) ...p( z_{d})}\right) dz_{1} ...dz_{d}
\end{align}

\subsection{Disentangled Representation Learning for ST data}
There is no prior research on disentangling ST data\cite{denton2017unsupervised}. Extracting information from both spatial and temporal domains is a challenge since the existing disentangled representation learning method is not designed for temporal behaviours. Therefore, it is needed to design a framework that can take account of both spatial and temporal domains and explore the downstream task utility.
\begin{figure*}[ht]
\begin{center}
    \includegraphics[width=0.7\textwidth]{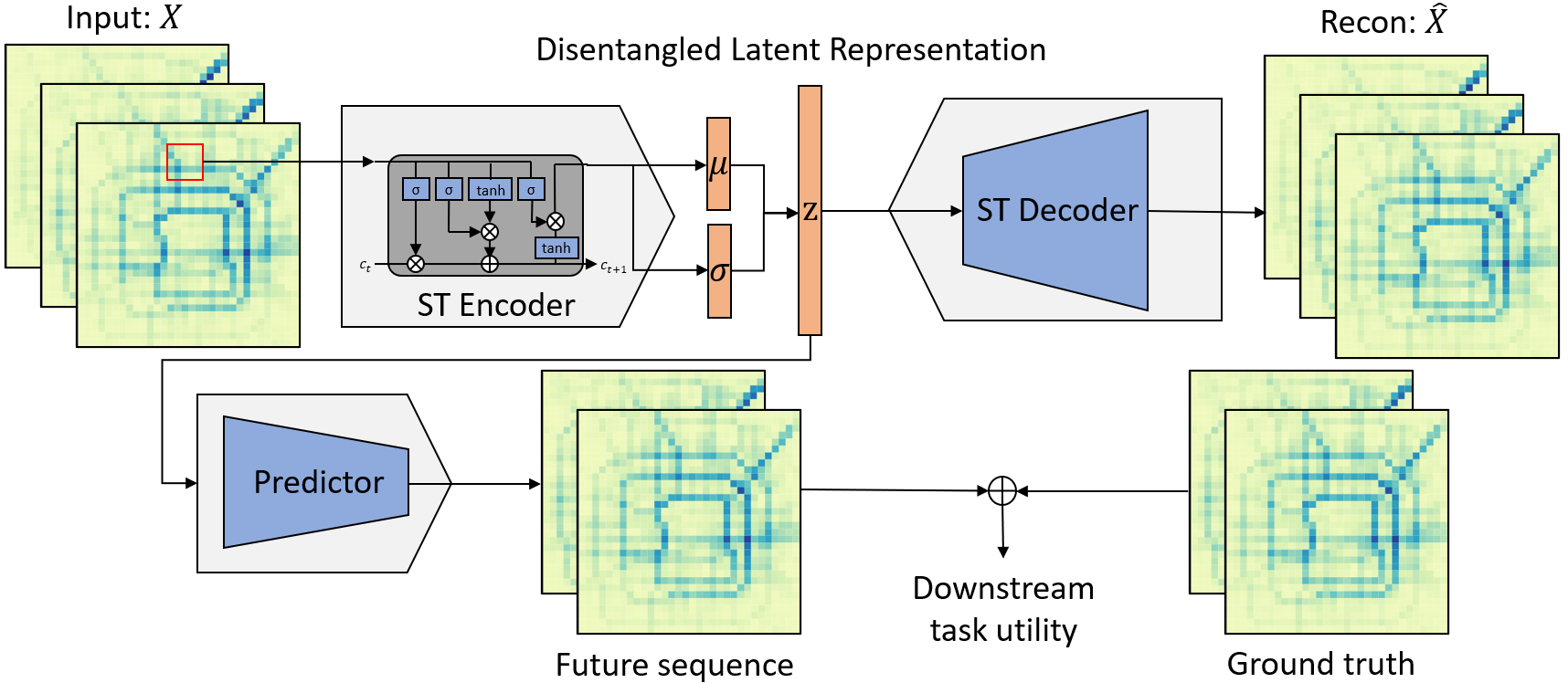}
\end{center}
\caption{ST-VAE: Our proposed framework for disentangled representation learning on ST data. The ST Encoder will extract spatio-temporal features in a convolutional manner over a period of time and then disentangle them using the FactorVAE/$\beta$-VAE's constraint.}
\label{arch}
\end{figure*}
In this work, we propose our framework, ST-VAE, which adapt from two state-of-the-art disentangled representation learning methods: \textbf{$\beta$-VAE} \cite{betavae} and \textbf{Factor-VAE} \cite{factorVAE}. The structure of ST-VAE is shown in Figure \ref{arch}. In comparison to the current state-of-the-art methods, the encoder and decoder are specifically designed for ST data which enables the model to extract both spatial and temporal features. Besides, the ST encoder and decoder can be easily swapped by new ST deep learning modules for different types of ST data. 

For spatio-temporal raster data, we incorporate Convolutional LSTM (ConvLSTM) \cite{convlstm} and spatio-temporal residual networks (ST-ResNet) \cite{taxibj}, separately as ST encoder to extract information from both domains. They are the state-of-the-art methods for the spatio-temporal sequence prediction problem. ConvLSTM \cite{convlstm}, which replaced the linear matrix operations with convolution operations, can capture spatio-temporal correlations. As for the ST-ResNet, the whole time axis is divided into three fragments, denoting recent time, near history and distant history. A residual network is used on each fragment to capture spatial information.
We also introduce models such as Spatio-Temporal Graph Convolutional Networks (STGCN) to handle spatio-temporal graph data. STGCN \cite{stgcn} uses the combination of the temporal and spatial graph convolution layer to extract spatial and temporal dependencies and use them to predict traffic for future timestep. Additionally, a predictor is added, which takes the learned representation as input to indicate the downstream task utility mentioned in Section \ref{sec:downstream}.

\section{Experiments}
\subsection{Datasets}
In this study, we consider three large real-world spatio-temporal datasets as follows. 

\textbf{TaxiBJ (ST Raster) \cite{taxibj}} contains four years of crowd flows in Beijing from the year 2013 to 2016. The whole city is divided into a 32 by 32 grid map, and crowd flows are aggregated every 30 minutes except public holidays. 

\textbf{BikeNYC (ST Graph) \cite{bikenyc}} is a bike usage data collected from New York City’s Citi Bike bicycle sharing service. Each records in this dataset represent a bike trip that endures more than one minute. In this work, we transform the data into the hourly inflow/outflow for each station.

\textbf{MelbPed (ST Graph) \cite{melbped}} is collected by a sensor network in Melbourne and managed by the City of Melbourne. The sensor network contains over 50 sensors located across the city, and each sensor collects an hourly pedestrian count.

\subsection{Experiment settings }
\subsubsection{Hyperparameters setting}
Inspired by \cite{locatello2019challenging}, we test these models under six different hyperparameter settings, and for each setting, we repeat the experiment 50 times with different random seeds. All other variables, like the structure of the neural network, will be consistent throughout the entire experiment. For each set of hyperparameters, we trained two different ST-VAE models with FactorVAE and $/beta$-VAE's objective function separately. 

\subsubsection{The choice of ST encoder and decoder}
As discussed in the previous section, the ST encoder and decoder can be easily swapped by new ST deep learning modules for different types of ST data. For ST raster data, we test our method using ST-ResNet \cite{taxibj} and ConvLSTM \cite{convlstm}. We will use 'ST-VAE+STResNet' and 'ST-VAE+ConvLSTM' to represent them in later sections. As for ST graph data, we incorporate DCRNN \cite{dcrnn} and STGCN \cite{stgcn} for encoder and decoder and will represent them as 'ST-VAE+DCRNN' and 'ST-VAE+STGCN' in later sections.

\subsubsection{Downstream tasks}
As for the downstream tasks discussed in later sections, we use the learned disentangled representations as input to train several Multi-Layer Perceptron (MLP) to make predictions.

\subsection{Key experimental results}


\begin{table*}[h]
    \centering
    \caption{The results of reconstruction loss and utility loss.}
    \label{tab:results}
    \begin{tabular}{c|c|cc|cc}
        \hline\hline 
         \multirow{2}{*}{Dataset} & \multirow{2}{*}{Models} & \multicolumn{2}{c|}{Recon Loss} & \multicolumn{2}{c}{Utiliy Loss} \\
        \cmidrule{3-6} 
           &   & Best & Average & Best & Average \\
        \hline 
         \multirow{3}{*}{\makecell[c]{TaxiBJ\\(ST Raster)}} & FactorVAE & 1787.42 & N/A & 129.61 & N/A \\
          & ST-VAE+ST-ResNet & 546.13 & 613.68 & 126.17 & 150.56 \\
          & \textbf{ST-VAE+ConvLSTM} & \textbf{417.64} & \textbf{592.10} & \textbf{115.44} & \textbf{150.29} \\
        \hline 
         \multirow{3}{*}{\makecell[c]{MelbPed\\(ST Graph)}} & FactorVAE & 3.89 & N/A & 869.46 & N/A \\
          & ST-VAE+DCRNN & 3.52 & 3.77 & 995.39 & 1010.83 \\
          & \textbf{ST-VAE+STGCN} & \textbf{3.13} & \textbf{3.69} & \textbf{857.04} & \textbf{945.21} \\
        \hline 
         \multirow{3}{*}{\makecell[c]{BikeNYC\\(ST Graph)}} & FactorVAE & 19.88 & N/A & 0.64 & N/A \\
          & ST-VAE+DCRNN  & 6.43 & 13.12 & 0.09 & 8.72 \\
          & \textbf{ST-VAE+STGCN} & \textbf{1.34} & \textbf{8.59} & \textbf{0.05} & \textbf{5.04} \\
         \hline\hline
    \end{tabular}
\end{table*}

\begin{figure*}[h]
\centering
  \subcaptionbox{Disentangled using FactorVAE constraint\label{mi_tc_factorvae}}{%
    \includegraphics[width=0.35\textwidth]{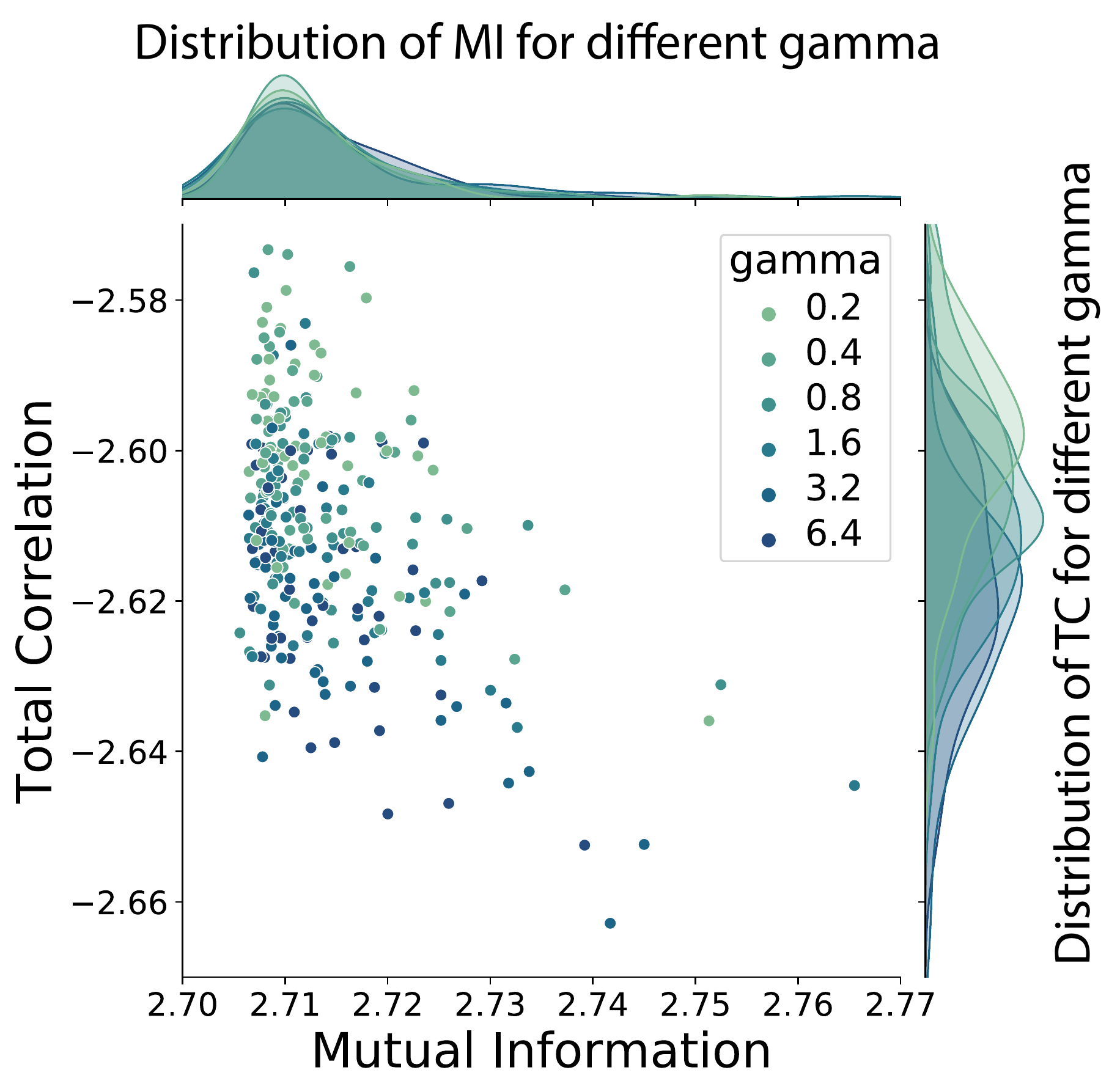}
  }
  \subcaptionbox{Disentangled using $\beta$-VAE constraint\label{mi_tc_betavae}}{%
    \includegraphics[width=0.35\textwidth]{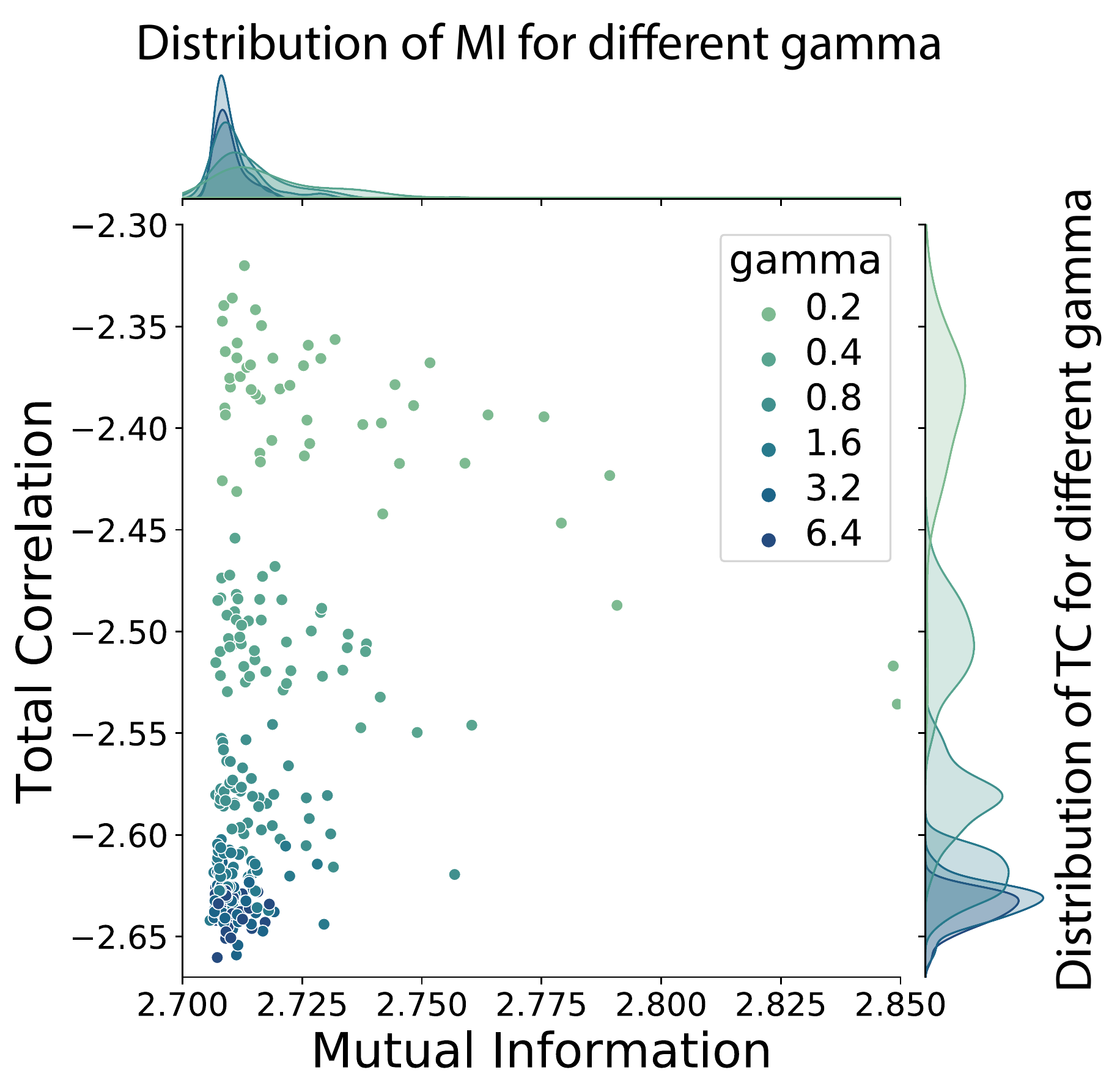}
  }
\caption{Correlation between the index-code mutual information and total correlation under different gamma values. The stronger constraints on regulariser show a impact on total correlation, but the distribution of the mutual information remains unchanged.}
\label{mi_tc_factorvae}
\end{figure*}

In this section, we summarise the experiment results and highlight our key findings with plots based on some problems we discussed in the previous sections.



\begin{figure*}[!h]
\centering
  \subcaptionbox{Disentanglement score Vs. Total Correlation\label{dis_tc_factorvae}}{%
    \includegraphics[width=0.35\textwidth]{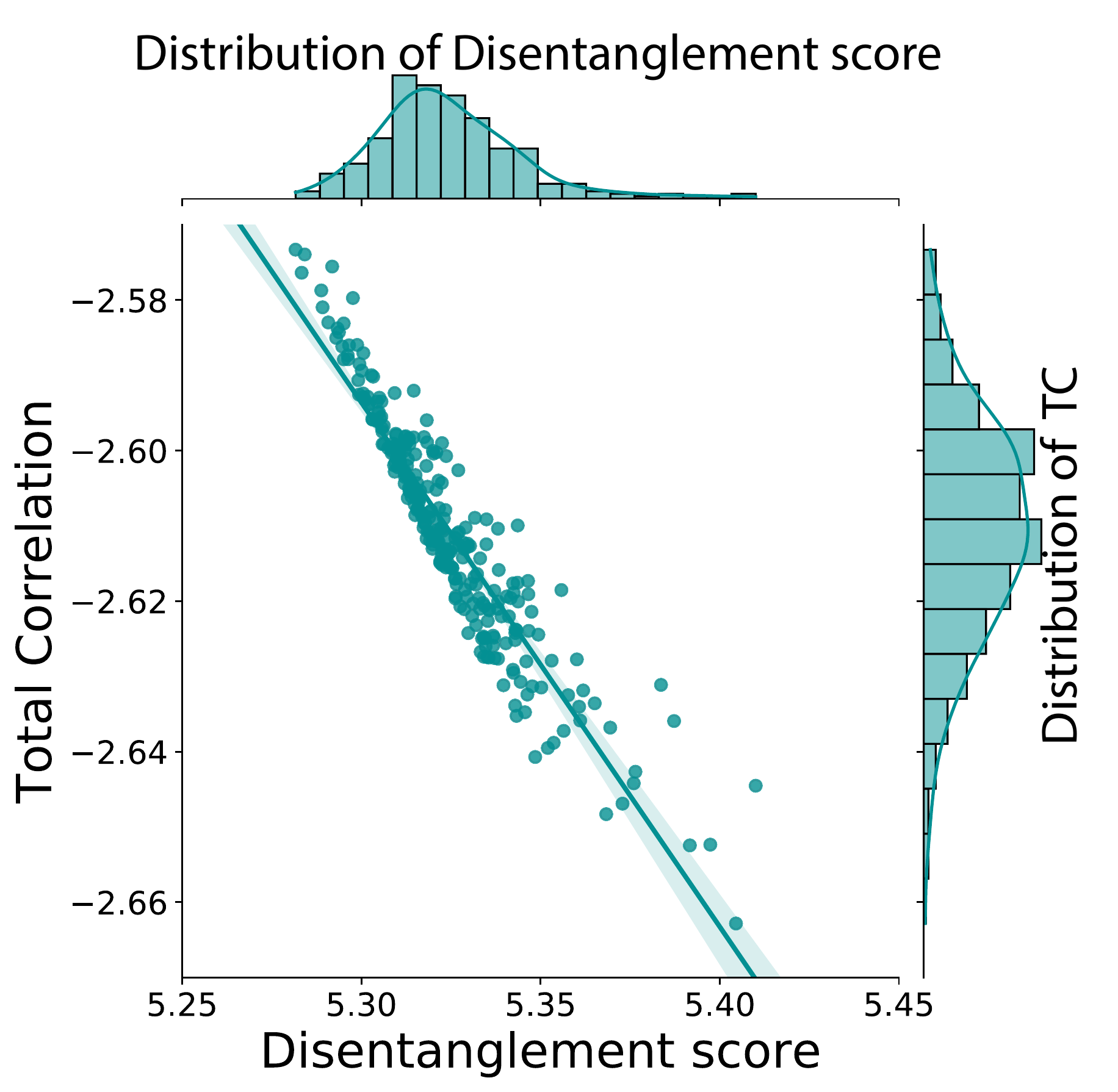}
  }
  \subcaptionbox{Disentanglement score Vs. Utility loss\label{dis_utility_factorvae}}{%
    \includegraphics[width=0.35\textwidth]{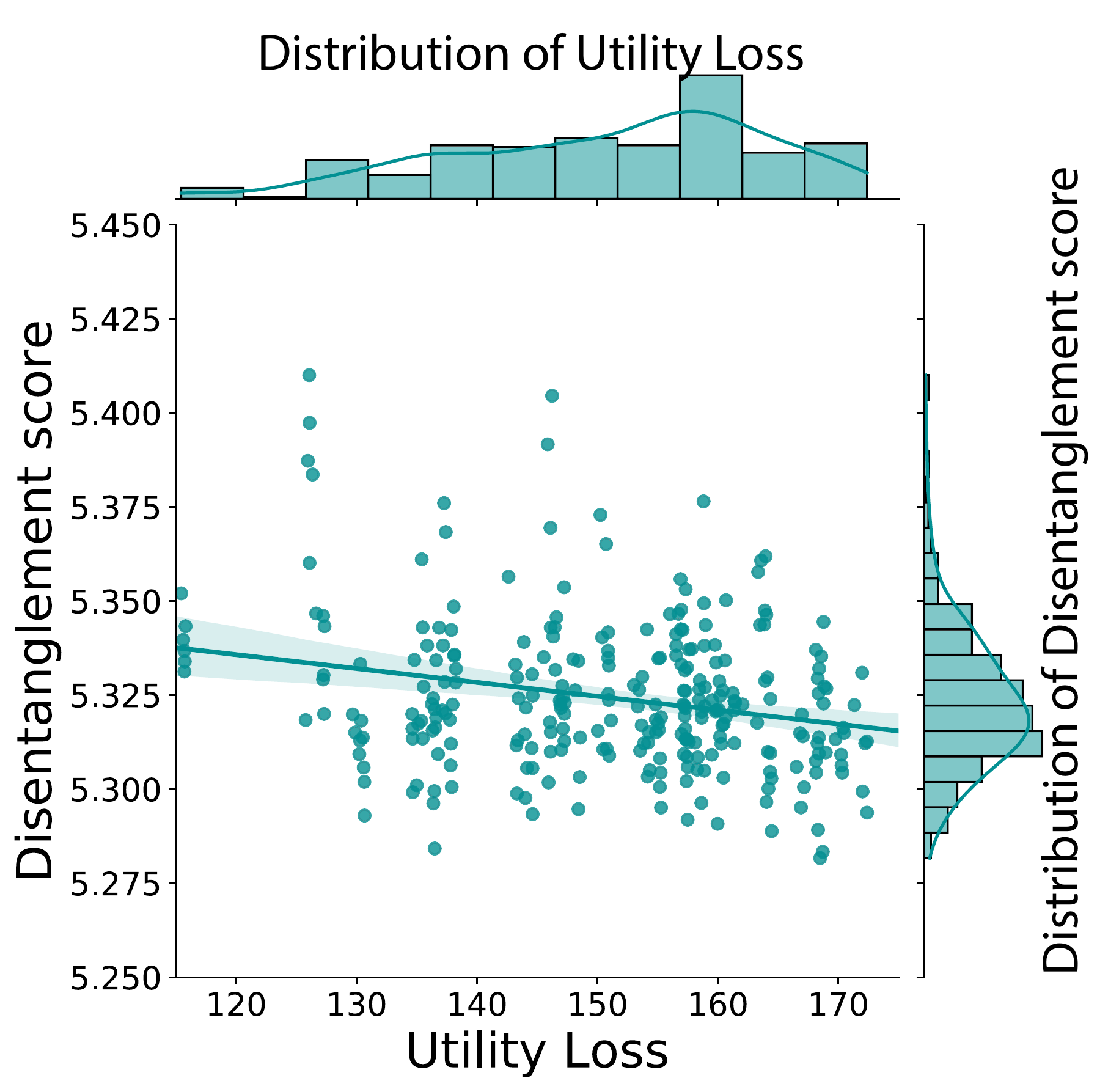}
  }
\caption{The correlation between Disentanglement Score and Total Correlation/Utility loss. (a) The diagonal line is a linear regression that shows a high correlation between TC and the disentanglement score. (b) There is no obvious correlation between the downstream task utility and disentanglement score.}
\label{img:dis}
\end{figure*}

\begin{figure*}[h]
  \centering
  \begin{center}
    \includegraphics[width=0.9\textwidth]{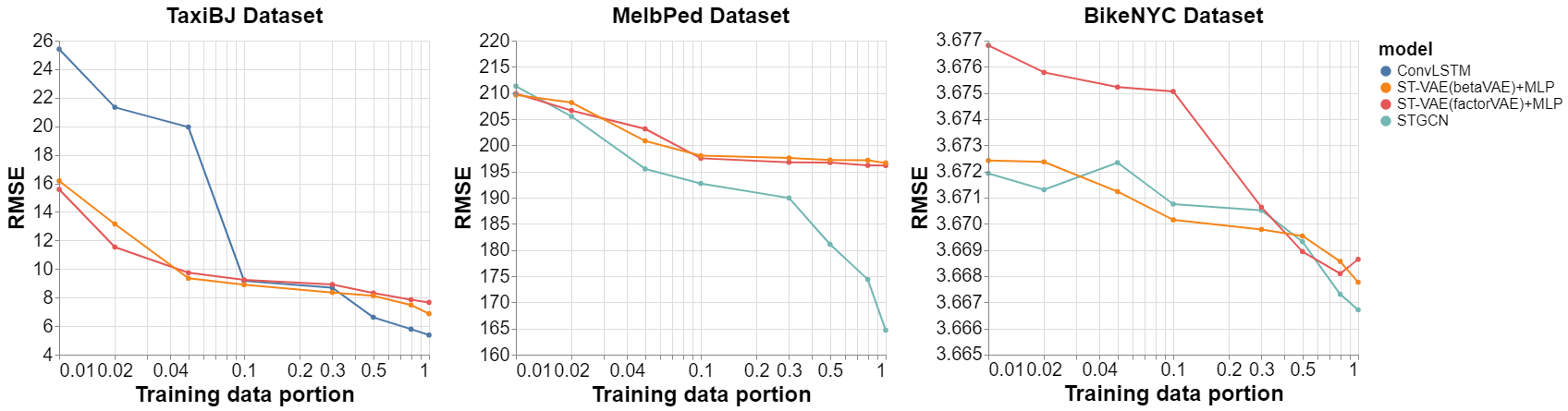}
    \end{center}
  \caption{The RMSE loss for models that are trained on different portion of training data (from 1\%to 100\%) on three ST datasets.}
  \label{fig:portion_results}
\end{figure*}

\subsubsection{How to learn a disentangled representation from spatio-temporal data.}
Compared to the existing methods typically used for image data, our proposed framework shows good results in extracting information from both spatial and temporal domains. 
As shown in Table \ref{tab:results}, our methods (ST-VAE) show better performance with respect to reconstruction loss and utility loss in the downstream task. Both modified ST models perform better than standard FactorVAE (treating ST raster as image data) indicates that the introduction of ST encoder and decoder enhance the models' ability to capture temporal correlations in the spatio-temporal data. However, the results of the BikeNYC dataset shows that there is still a large gap between different ST models, even for the same type of ST data. Therefore, there is no general ST module for all ST datasets at the time, and other methods' performance should be investigated.

\subsubsection{How to evaluate the quality of a disentangled representation of ST data.}
Some works suggest that putting constraints on the \textbf{Total Correlation} for a more disentangled representation \cite{burgess2018understanding,achille1706emergence}. However, the impact of higher constraints on disentangled representation learning of spatio-temporal data is unclear. 

In this section, we first look at the question, "Does the push of disentanglement compromises the model's ability to preserve information from its input?". Therefore we calculate the estimated mutual information and total correlation in Figure \ref{mi_tc_factorvae}. On the one hand, since the total correlation indicates the independence between latent variables, by applying stronger constraints (higher gamma value), we should observe the decrease of the total correlation. On the other hand, the mutual information states how much information about the input $x$ we have managed to preserve when mapping to $z$-space; there should not be obvious changes in the distribution of mutual information values. As shown in Figure \ref{mi_tc_factorvae}, the darker colour indicates a tighter bottleneck caused by a stronger constraint (higher gamma value). 
There is no obvious change for the distribution of mutual information under different gamma for both the factorVAE and $\beta$-VAE architectures, and the distribution of $TC(z)$ is shifting down when higher regularising strength is applied, which indicates that a higher penalty on $TC(z)$ encourages the model to get a more compact and disentangled representation, which matches the pattern mentioned by the state-of-the-art methods for image data. Most mutual information values remain on the same level, which states that the constraints will not affect the amount of information preserved by the learned latent variable. We found that for spatio-temporal data, the higher constraints on the total correlation can encourage the model to learn more disentangle representation while still managing to preserve enough information from the input.

Another question we want to discuss is, "Why to use the Mutual information value minus the Total correlation value as the disentanglement score?". We calculate the sum of mutual information $I(x,z)$ and total correlation $TC(z)$ and plot them Figure \ref{dis_tc_factorvae}. The intention here is to gain some insight about which learned representation should be used for the downstream task. A model with a high mutual information value and a low total correlation value should be used since it indicates that the learned representation has a high level of disentanglement while maintaining the ability to preserve information from its input. Figure \ref{dis_tc_factorvae} shows that the disentanglement score is highly correlated with the total correlation term. The value of the mutual information term only has a small variation even for $\beta$-VAE model, which emphasises the penalty on it by hyperparameter $\beta$. It is better to choose the model on the right bottom corner for your other tasks.

\subsubsection{The relationship between downstream tasks utility and disentanglement score}
The downstream task utility is another assumption that needs to be put to the test. There are some debates about whether the downstream tasks can benefit from a disentangled representation \cite{bengio2013representation,locatello2019challenging}, and it remains unknown in terms of spatio-temporal data. We perform two different experiments in this section. In the first experiment, we calculate the correlation between the RMSE loss of the simple task and the Total Correlation $TC(z)$ in Figure \ref{dis_utility_factorvae}. If encouraging disentanglement has a negative impact on the utility of the downstream task, we should observe a regression line that lies on the diagonal line with the dot spreads around it. However, in this work, we observed no strong correlations between the utility and the disentanglement level of the representation. The independence of the latent variables will not affect the downstream task utility. Therefore, we could obtain a model with both effective disentangled representation and good downstream utility with fine-tuning.

The second experiment aims to validate that the new representation is better than the raw data for prediction tasks. Firstly, we choose some spatio-temporal deep learning models as baselines and feed them with the raw data. We then train some simple Multi-Layer Perceptron (MLP) models using the latent representations learned by our ST-VAE as input. We repeat this process many times, and at each time, the models are trained with a different portion of training data (from 1$\%$ to 100$\%$). The model's task in this test is to predict more steps down the given sequence instead of only one in the simple task, and the results are plotted in Figure \ref{fig:portion_results}. Since the learned representation is in a much more compact format while still preserving enough information from the raw input, its performance should be better than just using the raw input. And in Figure \ref{fig:portion_results}, we observed that the task will benefit from the learned disentangled representation when the training portion is small. However, when the training portion gets larger, the advantage of that representation disappears and is overtaken by the baseline model. Although the disentangled representation may have advantages under a few-shot learning schema, we still need to interpret this with care since the mechanism leading to this result is still unclear.

\section{Conclusion}
This work discusses the challenges of applying disentangled representation learning to spatio-temporal data. Also, we proposed a novel metric to quantify and evaluate a disentangled generative representation aimed at both the disentanglement and utilities of the representation. To validate our claim, we perform empirical experiments on several large real-world spatio-temporal (ST) datasets, particularly ST graph and raster data. We draw a few conclusions from the experiments: 1) There is no guarantee either for spatio-temporal datasets that better disentanglement leads to lower downstream task loss. 2) The learned disentangled representation may have better predictive performance in a scenario when a small portion of input data is used in the training stage. For future work, we would like to explore the effectiveness of disentangled spatio-temporal representation in few-shot learning to investigate the performance of different ST modules further.


\begin{thebibliography}{99}
\bibitem{bengio2013representation}
Y.~Bengio, A.~Courville, and P.~Vincent, {\em Representation learning: A review and new perspectives}, IEEE transactions on pattern analysis and machine intelligence, 35 (2013), pp.~1798-1828.

\bibitem{su2018learning}
 B. Su and Y. Wu, {\em Learning low-dimensional temporal representations}, International Conference on Machine Learning, 2018: PMLR, pp. 4761-4770. 

\bibitem{betavae}
I. Higgins et al., {\em beta-vae: Learning basic visual concepts with a constrained variational framework }, 2016.

\bibitem{factorVAE}
H. Kim and A. Mnih, {\em Disentangling by factorising}, International Conference on Machine Learning, 2018: PMLR, pp. 2649-2658. 

\bibitem{kramer1991nonlinear}
M. A. Kramer, {\em Nonlinear principal component analysis using autoassociative neural networks}, AIChE J., 37 (1991), pp. 233-243, 1991.

\bibitem{Goodfellow-et-al-2016}
I. Goodfellow, Y. Bengio, and A. Courville, {\em Deep learning}, MIT press, 2016.

\bibitem{henaff2019data}
O. Henaff, {\em Data-efficient image recognition with contrastive predictive coding}, International Conference on Machine Learning, 2020: PMLR, pp. 4182-4192. 

\bibitem{creager2019flexibly}
 E. Creager et al., {\em Flexibly fair representation learning by disentanglement}, International conference on machine learning, 2019: PMLR, pp. 1436-1445. 

\bibitem{chen2018isolating}
R. T. Chen, X. Li, R. Grosse, and D. Duvenaud, {\em Isolating sources of disentanglement in variational autoencoders}, arXiv preprint arXiv:1802.04942, 2018.

\bibitem{taxibj}
 J. Zhang, Y. Zheng, and D. Qi, {\em Deep spatio-temporal residual networks for citywide crowd flows prediction}, Thirty-first AAAI conference on artificial intelligence, 2017. 

\bibitem{locatello2019challenging}
 F. Locatello et al., {\em Challenging common assumptions in the unsupervised learning of disentangled representations}, International conference on machine learning, 2019: PMLR, pp. 4114-4124. 

\bibitem{vae}
D. P. Kingma and M. Welling, {\em Auto-encoding variational bayes}, arXiv preprint arXiv:1312.6114, 2013.

\bibitem{multivae}
 D. Bouchacourt, R. Tomioka, and S. Nowozin, {\em Multi-level variational autoencoder: Learning disentangled representations from grouped observations}, Thirty-Second AAAI Conference on Artificial Intelligence, 2018. 

\bibitem{zhu2020s3vae}
 Y. Zhu, M. R. Min, A. Kadav, and H. P. Graf, {\em S3VAE: Self-supervised sequential VAE for representation disentanglement and data generation}, Proceedings of the IEEE/CVF Conference on Computer Vision and Pattern Recognition, 2020, pp. 6538-6547. 

\bibitem{li2018disentangled}
Y. Li and S. Mandt, {\em Disentangled sequential autoencoder}, arXiv preprint arXiv:1803.02991, 2018.

\bibitem{babaeizadeh2017stochastic}
M. Babaeizadeh, C. Finn, D. Erhan, R. H. Campbell, and S. Levine, {\em Stochastic variational video prediction}, arXiv preprint arXiv:1710.11252, 2017.

\bibitem{denton2017unsupervised}
E. Denton and V. Birodkar, {\em Unsupervised learning of disentangled representations from video}, arXiv preprint arXiv:1705.10915, 2017.

\bibitem{hsieh2018learning}
J.-T. Hsieh, B. Liu, D.-A. Huang, L. Fei-Fei, and J. C. Niebles, {\em Learning to decompose and disentangle representations for video prediction}, arXiv preprint arXiv:1806.04166, 2018.

\bibitem{alfonso2010optimization}
L. Alfonso, A. Lobbrecht, and R. Price, {\em Optimization of water level monitoring network in polder systems using information theory}, Water Resources Research, vol. 46, no. 12, 2010.

\bibitem{li2020pri}
Y. Li, S. Yu, J. C. Principe, X. Li, and D. Wu, {\em PRI-VAE: principle-of-Relevant-Information variational autoencoders}, arXiv preprint arXiv:2007.06503, 2020.


\bibitem{dsprites17}
L. Matthey, I. Higgins, D. Hassabis, and A. Lerchner, {\em dsprites: Disentanglement testing sprites dataset}, 2017.

\bibitem{liu2015faceattributes}
Z. Liu, P. Luo, X. Wang, and X. Tang, {\em Deep learning face attributes in the wild}, Proceedings of the IEEE international conference on computer vision, 2015, pp. 3730-3738. 

\bibitem{aubry2014seeing}
 M. Aubry, D. Maturana, A. A. Efros, B. C. Russell, and J. Sivic, {\em Seeing 3d chairs: exemplar part-based 2d-3d alignment using a large dataset of cad models}, Proceedings of the IEEE conference on computer vision and pattern recognition, 2014, pp. 3762-3769. 

\bibitem{achille1706emergence}
A. Achille and S. Soatto, {\em On the emergence of invariance and disentangling in deep representations}, arXiv preprint arXiv:1706.01350, vol. 125, pp. 126-127, 2017.

\bibitem{convlstm}
 S. Xingjian, Z. Chen, H. Wang, D.-Y. Yeung, W.-K. Wong, and W.-c. Woo, {\em Convolutional LSTM network: A machine learning approach for precipitation nowcasting}, Advances in neural information processing systems, 2015, pp. 802-810. 

\bibitem{stgcn}
B. Yu, H. Yin, and Z. Zhu, {\em Spatio-temporal graph convolutional networks: A deep learning framework for traffic forecasting}, arXiv preprint arXiv:1709.04875, 2017.

\bibitem{bikenyc}
L. a. J. C. B. S. NYC Bike Share, LLC. {\em Citi Bike System Data}, https://www.citibikenyc.com/system-data (accessed 2021).

\bibitem{melbped}
X. Wang, J. Liono, W. Mcintosh, and F. D. Salim, {\em Predicting the city foot traffic with pedestrian sensor data}, Proceedings of the 14th EAI International Conference on Mobile and Ubiquitous Systems: Computing, Networking and Services, 2017, pp. 1-10. 

\bibitem{burgess2018understanding}
C. P. Burgess et al., {\em Understanding disentangling in $\beta $-VAE}, arXiv preprint arXiv:1804.03599, 2018.

\bibitem{dcrnn}
Y. Li, R. Yu, C. Shahabi, and Y. Liu, {\em Diffusion convolutional recurrent neural network: Data-driven traffic forecasting}, arXiv preprint arXiv:1707.01926, 2017.













\end{thebibliography}
\end{document}